\RequirePackage{fix-cm}
\documentclass[twocolumn]{svjour3}          % twocolumn
\smartqed  % flush right qed marks, e.g. at end of proof
\usepackage{graphicx}
\usepackage{wrapfig}
\usepackage{color}

\usepackage{float}
\usepackage{amsmath}
\usepackage{amssymb}
\usepackage[sort,numbers]{natbib}
\usepackage[a4paper, total={6in, 8in}]{geometry}

\begin{document}
% For all papers calls, please submit 8-12 pages.
% \begin{sloppypar}

\title{Segment Any Building For Remote Sensing}
% \subtitle{Segment Any Building}
\author{Lei Li}
\institute{Computer Science Department of Copenhagen University, lilei@di.ku.dk}
\date{} % The correct dates will be entered by the editor

\maketitle

\begin{abstract}
The task of identifying and segmenting buildings within remote sensing imagery has perennially stood at the forefront of scholarly investigations. This manuscript accentuates the potency of harnessing diversified datasets in tandem with cutting-edge representation learning paradigms for building segmentation in such images. Through the strategic amalgamation of disparate datasets, we have not only expanded the informational horizon accessible for model training but also manifested unparalleled performance metrics across multiple datasets. Our avant-garde joint training regimen underscores the merit of our approach, bearing significant implications in pivotal domains such as urban infrastructural development, disaster mitigation strategies, and ecological surveillance. Our methodology, predicated upon the fusion of datasets and gleaning insights from pre-trained models, carves a new benchmark in the annals of building segmentation endeavors. The outcomes of this research both fortify the foundations for ensuing scholarly pursuits and presage a horizon replete with innovative applications in the discipline of building segmentation.

\end{abstract}

% polish and expand for one paragraph with more academic

\section{Introduction}
\label{sec:1}
The building environment, constituting a diverse spectrum of structures, remains a crucial facet of our urban and rural landscapes. Structures, ranging from residential and commercial spaces to industrial facilities, play instrumental roles in shaping economic dynamics, facilitating societal interactions, and influencing environmental outcomes. Consequently, the task of building segmentation and subsequent analysis holds paramount significance across an array of disciplines, including but not limited to urban planning, real estate, and disaster management ~\cite{rahnemoonfar2021floodnet}. These analytical processes provide indispensable insights and contribute to both the theoretical understanding and practical applications within these domains, affirming the necessity of building segmentation in contemporary academic research and industry practice.

Building segmentation significantly depends on data derived from an array of imaging sources, chiefly encompassing high-resolution aerial photography and remote sensing imagery. Each of these sources offers unique vantage points and insights, which collectively contribute to a holistic understanding of built environments and forest management ~\cite{oehmcke2022deep,boguszewski2021landcover,revenga2022above}. High-resolution aerial photography, for instance, is instrumental in providing intricately detailed depictions of buildings and their immediate surroundings. These close-up views are invaluable for conducting fine-grained analyses that delve into the minutiae of individual structures and their architectural features. On the other hand, remote sensing imagery affords a more macroscopic perspective, capturing expansive urban and rural areas. The broader view provided by such imagery facilitates large-scale analyses and enables comparative studies across extensive geographical regions. Together, these data sources, each with its own strengths, enrich the process of building segmentation by offering different layers of information, ultimately allowing researchers to unearth nuanced understandings of the built environment from multiple scales and perspectives.

Despite the valuable insights provided by high-resolution aerial photography and satellite imagery ~\cite{oehmcke:22}, these data sources do present inherent challenges that must be acknowledged. The primary limitation of aerial photography lies in its restricted spatial coverage, rendering it less applicable for expansive geographical analyses. In contrast, satellite imagery, while boasting extensive coverage, often suffers from a relatively lower resolution, potentially compromising the detail of analytical outputs. Both are also susceptible to image quality inconsistencies due to varying atmospheric conditions during data acquisition. Furthermore, significant differences may arise in their optical properties due to variations in camera technologies and configurations, which can impact image colorimetry, contrast, and sharpness. These discrepancies emphasize the need for sophisticated calibration methods and advanced image processing techniques to mitigate potential inaccuracies and maximize the utility of each data source in the field of building segmentation.

The distinct characteristics of high-resolution aerial photography and satellite imagery necessitate tailored methodological approaches for each data source. This requirement emerges from variances in resolution, camera properties, and imaging conditions, which imply that analytical techniques successful with one data type may not achieve equivalent accuracy with another. It's imperative to identify and address these challenges to push forward in building segmentation research and its numerous applications. Understanding these complexities can lead to the development of robust, source-specific analytical models, capable of harmonizing data from varied sources to enhance building segmentation accuracy. This approach can potentially lead to innovative breakthroughs in related fields like urban planning, environmental monitoring, and disaster management.

Acknowledging the challenges inherent to the differing data sources, and building upon the recent advancements in the field of general segmentation, particularly the Segment Anything (SA) ~\cite{kirillov2023segment} method, we adopt a nuanced approach. Our strategy involves the amalgamation of multiple datasets processed through pretrained models, thereby addressing data discrepancies and facilitating mutual learning across various data domains.

Our research contributions within the realm of building segmentation are manifold:

Firstly, we harness the robust framework of the SSA method, utilizing its capacity for extensive data processing within large models. This aids in augmenting the precision and efficiency of building segmentation tasks, illustrating the value of integrating sophisticated algorithms within such expansive data processes.

Secondly, we confront the issue of inter-data discrepancies through the individual processing of various datasets, thereby encouraging cross-domain learning. This approach not only serves to alleviate the constraints tied to individual data sources but also amplifies the overall learning process through the incorporation of diverse and extensive information.

Lastly, our adapted method exhibits commendable results across an array of datasets, thereby underlining its efficacy and flexibility. This superior performance, regardless of dataset variability, reinforces the potential of our approach in providing general insights within the field of building segmentation.

This article will proceed as follows: an exploration of the related work in the field forms the initial focus, providing a contextual understanding of the current state of building segmentation research. This is followed by an in-depth examination of the specific methods employed in our approach and a comprehensive discussion of the corresponding data performance. Subsequently, we present the experimental results, augmented by visualizations to provide a tangible representation of our findings. The article culminates with discussion and conclusion that encapsulates the core insights derived from our research and their implications for future work in the field.

\section{Related work}
\label{sec:2}
\paragraph{Image Segmentation.} 
Image Segmentation, as an essential step in image analysis and interpretation, has received considerable attention in academic research over the past few years. Such as Unet~\cite{ji2018fully}, Segformer~\cite{xie2021segformer}, Deeplab~\cite{chen2017deeplab}, ConNext ~\cite{liu2022convnet}. The body of work spans across various techniques and methods, ranging from traditional threshold-based and region-growing methods to more advanced machine learning and deep learning techniques. Some work ~\cite{li2022buildseg} In particular, the U-Net architecture, first introduced by Ronneberger et al ~\cite{ronneberger2015u}. in 2015, has been widely adopted for biomedical image segmentation due to its impressive performance and then been utilized to other different data domains. However, despite the strides made in this field, segmentation remains a challenging task due to issues such as the variability of object shapes and sizes, background clutter, and imaging conditions, thus, necessitating continued exploration and innovation in this area.
\paragraph{Image Data fusion.} Image data fusion has emerged as a critical process in numerous image segmentation tasks~\cite{zhang2023attention,cheng2022masked,li2023mask, he2017mask}, including building segmentation, due to its capacity to combine complementary information from multiple data sources, thereby enhancing the quality and utility of the resulting data. Extensive literature exists concerning the methods and techniques utilized for this purpose. Conventional methods often incorporate mathematical transformations such as Principle Component Analysis (PCA) ~\cite{zhang2022lr, arbelaez2010contour, zhang2023attention,li2023edge} and Intensity Hue Saturation (IHS) for fusing low-resolution multispectral data with high-resolution panchromatic data. Recent years have witnessed a surge in research exploring machine learning and deep learning techniques for image fusion. Channel fusion ~\cite{zhang2020fact}, particularly Convolutional Neural Networks (CNNs), have been employed for their ability to learn complex and high-level features from multi-source data with similarity ~\cite{wu2019fase}. Despite the significant advancements in this domain, the fusion of image data remains a nontrivial task due to issues like preserving spectral and spatial information and mitigating artifacts in fused images. This ongoing challenge underscores the need for continued research and development of sophisticated fusion techniques tailored to specific segmentation tasks.

\paragraph{Pre-trained model.}
The use of large pre-trained models ~\cite{bao2021beit} as the basis for various specialized tasks is a prevalent strategy in the contemporary machine learning landscape. This approach leverages the broad feature learning capabilities of these models, which have been trained on extensive and diverse datasets, thus providing a robust starting point for a variety of specialized tasks. Models such as BERT~\cite{devlin2018bert}, GPT, and SA ~\cite{kirillov2023segany}, for instance, have shown significant efficacy when fine-tuned for specific tasks like text classification, object detection, and semantic segmentation, amongst others. These models offer the advantage of leveraging transfer learning, which allows the application of learned features to new tasks, thereby reducing the need for extensive data and computational resources.

In the context of image segmentation tasks, the use of pre-trained models has been increasingly popular. For example, researchers have also investigated the use of large pre-trained models like VGG16 and VGG19 ~\cite{simonyan2014very} for tasks like building segmentation. These models are typically fine-tuned on task-specific data, thus allowing the model to adapt its learned features to the unique characteristics of the new task.  Moreover, the use of prompt pre-trained Segment Anything(SA) ~\cite{kirillov2023segment} for semantic segmentation has been explored recently in several studies. While this approach has yielded promising results, the adaptation of large pre-trained models to new tasks is an ongoing area of research, with ample scope for exploring novel fine-tuning strategies and model architectures.

\section{Methods}
\label{sec:3}

\subsection{Problem formulation}
\begin{figure}[h]
    \centering
    \includegraphics[width=\columnwidth]{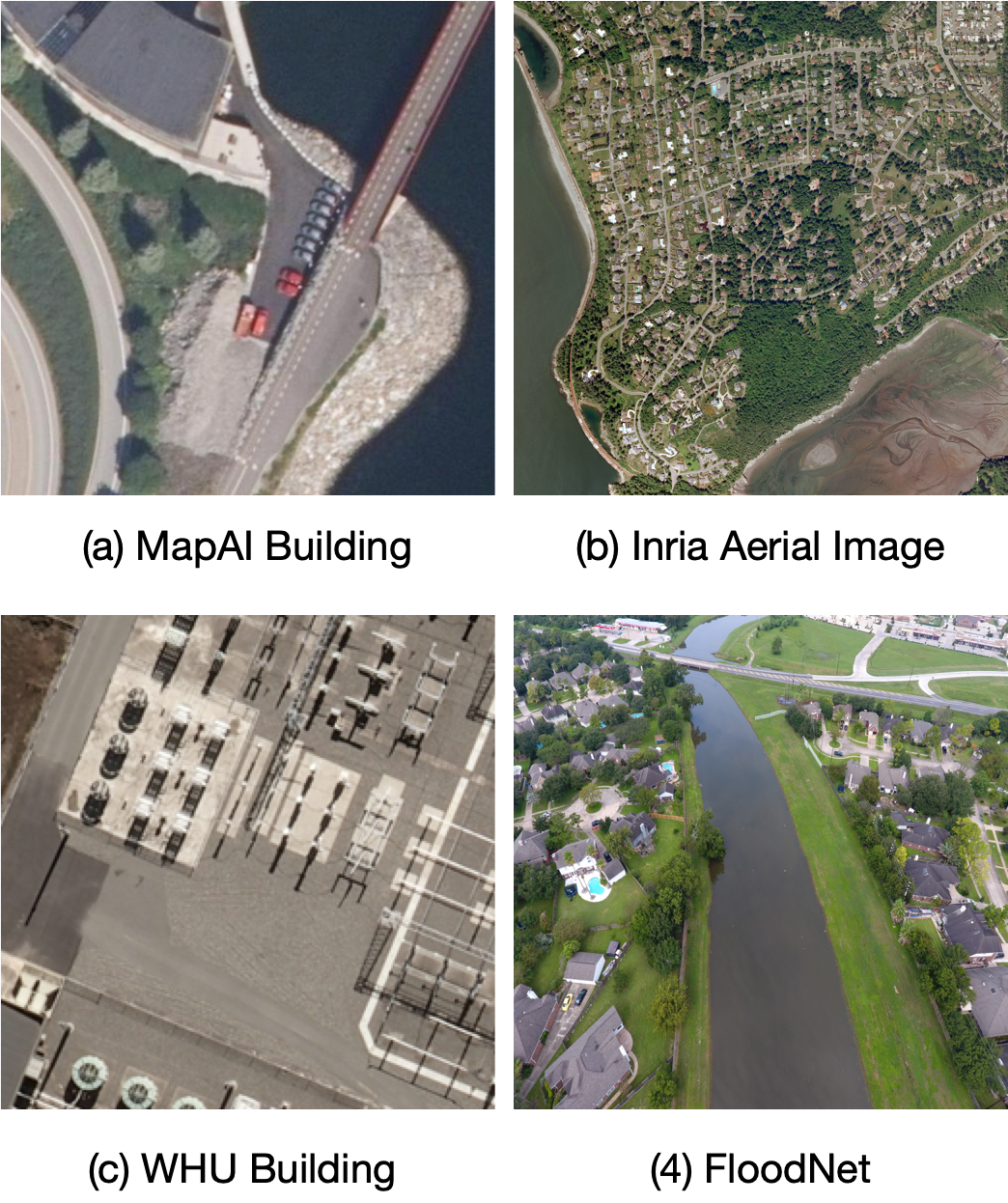}
    \caption{This study utilizes four distinct datasets, each embodying unique areas and scenes, non-intersecting in context, and are most effectively visualized through color view.}
    \label{fig:problem_formulation}
\end{figure}

In the Figure \ref{fig:problem_formulation}, The four datasets highlighted in this research, namely MapAI Building \cite{jyhne2022mapai}, Inria Aerial Image Labeling Benchmark\cite{maggiori2017dataset}, WHU Building Dataset \cite{ji2018fully}, and FloodNet\cite{rahnemoonfar2021floodnet}, each bring their unique strengths and nuances to building segmentation tasks. The MapAI Building dataset stands out for its incorporation of laser data and ground truth masks, along with aerial images. It covers diverse building types and environments spanning Denmark and Norway, and the real-world derived data poses unique challenges and authenticity. The Inria Aerial Image Labeling Benchmark excels in its wide geographical coverage and high-resolution aerial imagery. It uniquely tests the generalization capabilities of segmentation techniques across different regions, illumination conditions, and seasons. The WHU Building Dataset, comprising both aerial and satellite imagery, provides a comprehensive depiction across varied scales, geographical locations, and imaging sources. The segmentation task is further complicated by the presence of diverse remote sensing platforms. Lastly, the FloodNet dataset is specialized for disaster management with its UAS-based high-resolution imageries. The dataset uniquely categorizes building and road structures based on their flood status, thus bringing a novel dimension to building segmentation tasks. Each dataset's idiosyncrasies underscore the need for adaptable segmentation methods capable of handling different data types, resolutions, and scenario-specific complexities.

\subsection{Data}

\paragraph{MapAI Building\cite{jyhne2022mapai}.} The dataset employed in this research amalgamates aerial images, laser data, and ground truth masks corresponding to building structures, catering to a diverse range of environmental and building types. The training dataset is composed of data derived from multiple locations across Denmark, thereby ensuring considerable variability and diversity in the nature of the data. Conversely, the test dataset consists of seven distinct locations in Norway, encompassing both urban and rural environments. It's worth noting that the data originates from real-world scenarios, leading to certain instances where buildings in the aerial images do not align with the corresponding ground truth masks. An additional complexity is the method of generating ground truths in the test dataset using a Digital Terrain Model (DTM), which results in a certain degree of skewness in the building tops in the images, compared to the ground truths. In contrast, the training dataset ground truths are generated using a Digital Surface Model (DSM), which effectively circumvents the issue of skewness in the building tops. The full dataset will be released following the competition, thus enabling further examination and research.

\paragraph{Inria Aerial Image Labeling Benchmark \cite{maggiori2017dataset}.} 
The dataset under investigation is characterized by extensive coverage, spanning 810 km², which is equally divided for training and testing purposes. It utilizes high-resolution (0.3 m) aerial orthorectified color imagery, which encompasses varied urban landscapes, from densely populated areas like San Francisco’s financial district to less dense regions like Lienz in Austrian Tyrol. The ground truth data comprises two semantic classes: 'building' and 'not building', publicly accessible only for the training subset. Unique to this dataset is its geographical division across training and testing subsets; training employs imagery from cities like Chicago, while testing uses data from different regions. This structure tests the techniques' generalization capabilities under diverse conditions including varied illumination, urban landscape, and seasons. The dataset's assembly involved merging public domain imagery and official building footprints, providing a comprehensive depiction of building structures.

\paragraph{WHU Building Dataset \cite{ji2018fully}.} The WHU building dataset, meticulously curated for this study, incorporates both aerial and satellite imagery of building samples. The aerial component of the dataset comprises over 220,000 distinct building structures, gleaned from aerial images with a fine spatial resolution of 0.075 m, and spans an area of 450 km² in Christchurch, New Zealand. The satellite imagery dataset is bifurcated into two subsets: one encompasses images from diverse cities globally, sourced from multiple remote sensing platforms including QuickBird, Worldview series, IKONOS, ZY-3, among others, encapsulating a broad range of geographic and urban contexts. The second subset consists of six contiguous satellite images covering an expanse of 550 km² in East Asia with a ground resolution of 2.7 m. Collectively, the WHU building dataset offers a comprehensive and varied collection of images, affording the opportunity to explore building segmentation across different scales, geographical locations, and imaging sources.

\paragraph{Floodnet \cite{rahnemoonfar2021floodnet}.} The FloodNet dataset is a meticulously curated resource aimed at revolutionizing disaster management through the provision of high-resolution and semantically detailed unmanned aerial system (UAS) imagery, specifically in the context of natural disasters such as hurricanes. It leverages the flexible and efficient data collection capabilities of small UAS platforms, namely DJI Mavic Pro quadcopters, which are especially valuable for rapid response and recovery in large-scale and difficult-to-access areas. The dataset was collated in the aftermath of Hurricane Harvey and comprises 2343 images, apportioned into training (approximately 60\%), validation (around 20\%), and testing (roughly 20\%) subsets. The semantic segmentation labels within the dataset are notably comprehensive, covering categories such as background, flooded and non-flooded buildings, flooded and non-flooded roads, water bodies, trees, vehicles, pools, and grass. Despite the wealth of data provided by such UAS platforms, analyzing these large datasets and extracting meaningful information presents a considerable challenge, underscoring the significance of FloodNet's detailed semantic annotation in advancing disaster management research and applications.

\subsection{Overview}
\begin{figure*}[t]
    \centering
    \includegraphics[width=\textwidth]{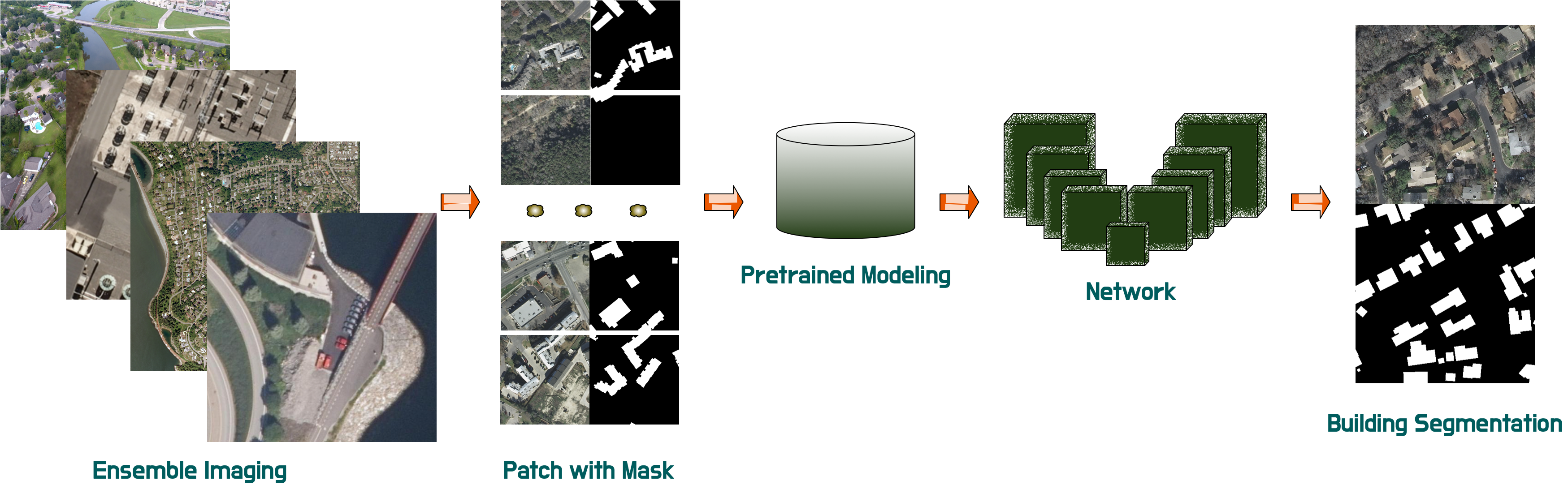}
    \caption{\textbf{The propose SegAnyBuild framework.} Initially, we align the architectural structures and corresponding images as per the utilized dataset. Subsequently, the entire dataset is homogenized into a uniform 256*256 patch accompanied by a mask. Feature pre-learning is performed utilizing the pretrained model of Semantic Segmentation Anything (SSA). Ultimately, the architectural structure is delineated via a segmentation network. The entire procedure is executed in an end-to-end manner.}
    \label{fig:network}
\end{figure*}

In our comprehensive methodology, we integrate four distinct datasets: MapAI Building, Inria Aerial Image Labeling Benchmark, WHU Building Dataset, and Floodnet. Irrespective of the original pixel resolution disparities, all data are reformatted into a standardized size of 256x256 pixels. Correspondingly, we apply similar alterations to the associated masks, enabling the generation of uniformly dimensioned image-mask patches.

We employ Segformer-B5 \cite{xie2021segformer} as our backbone, a decision underpinned by its robust performance in diverse segmentation tasks. To augment the model's initial capabilities, we incorporate pre-trained parameters, which are instrumental in defining the initial weights of our network, thereby optimizing our model's learning trajectory.

The culmination of our pipeline involves an up-sampling procedure, through which we transform the processed data into a style compatible with the target results. This rigorous, systematic approach aids in ensuring the efficacy of our segmentation processes and ultimately contributes to the production of efficient outputs.

\subsection{Network}

In this paragraph, we briefly introduce the more popular U-Net structure and the SegFormer network architecture. For more details about ConvNext, refer to ConvNext \cite{ConvNeXt}.

The U-Net architecture is a widely recognized model for biomedical image segmentation, characterized by its symmetric expansive and contracting structure. The architecture mathematically operates as a series of nonlinear mappings that progressively transform the input image into the output segmentation. Incorporating sequences of convolution operations, max-pooling, upsampling, and a softmax layer, this model is recognized for its effectiveness in detail retention and accurate segmentation.

The SegFormer architecture is an innovative blend of transformer and U-Net components, offering a novel approach to semantic segmentation tasks. A core component of the transformer section of the architecture is the self-attention mechanism, which can be represented mathematically as:

\begin{equation}
\text{Attention}(Q, K, V) = \text{softmax}\left(\frac{QK^T}{\sqrt{d_k}}\right)V
\end{equation}

In the given formulation, the variables \( Q \), \( K \), and \( V \) denote the query, key, and value vectors, respectively, which are extrapolated from the input feature mappings. The parameter \( d_k \) characterizes the dimensionality of the key vectors. Utilizing the \textit{softmax} operation ensures the normalization of these weights, mandating that their cumulative sum converges to unity. The term \(\frac{1}{\sqrt{d_k}}\) serves as a scaling factor, which is indispensable in ensuring a stable learning trajectory.

The \textit{SegFormer} model adopts the self-attention mechanism at heterogeneous scales, a concept more broadly recognized as multi-head attention. This paradigm is instrumental in amalgamating information from a diverse range of feature scales and can be succinctly delineated mathematically as:

\begin{equation}
\begin{split}
\text{MultiHead}(Q, K, V) = \text{Concat}(\text{head}_1,\\ \text{head}_2, ..., \text{head}_n)W_O
\end{split}
\end{equation}

Through this mechanism, the model aspires to achieve a nuanced comprehension of the encompassing input features by seamlessly integrating information across multiple scales.

In this equation, the output from the multi-head attention is subsequently integrated into a feature map, which is processed by a segmentation head to yield the final semantic segmentation output. ${head}_i = \\ \text{Attention}(QW_{Qi}, KW_{Ki}, VW_{Vi})$, with $W_{Qi}$, $W_{Ki}$, and $W_{Vi}$ being parameter matrices. $W_O$ serves as the output transformation matrix, while "Concat" represents the concatenation operation. 

\subsection{Loss function}
In the context of semantic segmentation tasks, the CrossEntropyLoss is often utilized to measure the dissimilarity between the predicted pixel-wise class probabilities and the ground truth labeling. This loss function is crucial for training the segmentation models as it encourages the accurate prediction of the class of each pixel in the image, for a single pixel, the cross entropy loss is defined as:

\begin{equation}
L(y, \hat{y}) = - \sum_{c=1}^C y_c \cdot \log(\hat{y}_c)
\end{equation}

Here, $C$ is the total number of classes, $y_c$ represents the ground truth (which would be 1 for the true class, and 0 for all other classes), and $\hat{y}_c$ denotes the predicted probability of the pixel belonging to class $c$. In our experiment, the classification task was binary, concentrating on differentiating between buildings and non-buildings.

\section{Experiments}
\label{sec:4}

\subsection{Metrics}
\paragraph{IOU}
Intersection over Union (IoU), also known as the Jaccard index, is a commonly utilized metric for the quantitative evaluation of segmentation tasks. Mathematically, it is defined as the ratio of the area of intersection to the area of union between the predicted and ground truth segmentation masks.

\begin{equation}
\text{IoU} = \frac{\text{Area of Intersection}}{\text{Area of Union}} = \frac{|A \cap B|}{|A \cup B|}
\end{equation}

In this equation, $A$ denotes the set of pixels in the predicted segmentation and $B$ represents the set of pixels in the ground truth. A higher IoU indicates a greater overlap between the predicted and actual regions, implying a more accurate segmentation.

\paragraph{BIOU}

We also use The Boundary Intersection over Union (BIoU) ~\cite{cheng2021boundary} as a important metric, extends the concept of traditional IoU by placing emphasis on the boundary pixels of the segmentation mask. The metric quantifies the degree of overlap between the boundaries of the predicted and ground truth segmentation masks.

\begin{equation}
\begin{split}
\text{BIoU} &= \frac{\text{Length of Intersection of Boundaries}}{\text{Length of Union of Boundaries}} \\
&= \frac{|B_A \cap B_B|}{|B_A \cup B_B|}
\end{split}
\end{equation}

In this equation, $B_A$ denotes the set of boundary pixels in the predicted segmentation and $B_B$ represents the set of boundary pixels in the ground truth. A higher BIoU signifies that the predicted and actual boundaries align more closely, signifying a more accurate delineation of the object's contours.

\subsection{Setting}
We executed a rigorous comparative study between the network configurations, delineating the performances with and without the integration of a pre-trained model. The results unequivocally demonstrated that the incorporation of a pre-trained model significantly enhances the network's performance, corroborating its strategic value in model optimization.

In the empirical analysis, we utilized the robust MMsegmentation
% ~\cite{mmseg2020} 
framework, conducting experiments on an advanced NVIDIA A100 Tensor Core GPU machine. We chose the SegFormer model for our examination, applying the AdamW optimizer. The initial learning rate was set to 0.0006, betas were in the range of 0.9 to 0.999, and the weight decay was adjusted to 0.01. We used a dynamic learning rate strategy, following a polynomial updating policy with a linear warmup phase of 1500 iterations. The warmup ratio was a minute 1e-6, and the power for the policy was set at 1.0, with a minimum learning rate of 0.0. These adjustments occurred within each iteration rather than at the epoch level. We carried out the experiments with a batch size of $32$. 

The backbone is a 'MixVisionTransformer', a unique form of Vision Transformer architecture that employs multi-scale inputs. It has four stages, each stage consisting of a different number of layers and attention heads. The input image will be divided into patches of varying sizes at different stages, with the size of the patches decreasing as we go deeper into the model, thanks to the patch sizes configuration. These patch sizes are related to the spatial reduction  at each stage. The model allows for a progressive decrease in spatial dimensions as we proceed from lower to higher stages.

The second part, the decode head, is a 'SegformerHead'. It takes the outputs from various stages of the backbone as inputs,  The number of input channels for each stage's output is specified in in channels. It then outputs a tensor with the number of channels equal to channels, where each channel corresponds to the prediction map of a specific class. The model uses a CrossEntropyLoss for the loss function during training. The align corners parameter is set to False to prevent the misalignment of corners in the bilinear interpolation during upsampling.

\subsection{Results}
\begin{figure*}[h!]
	\begin{center}
			\includegraphics[width=0.9\textwidth]{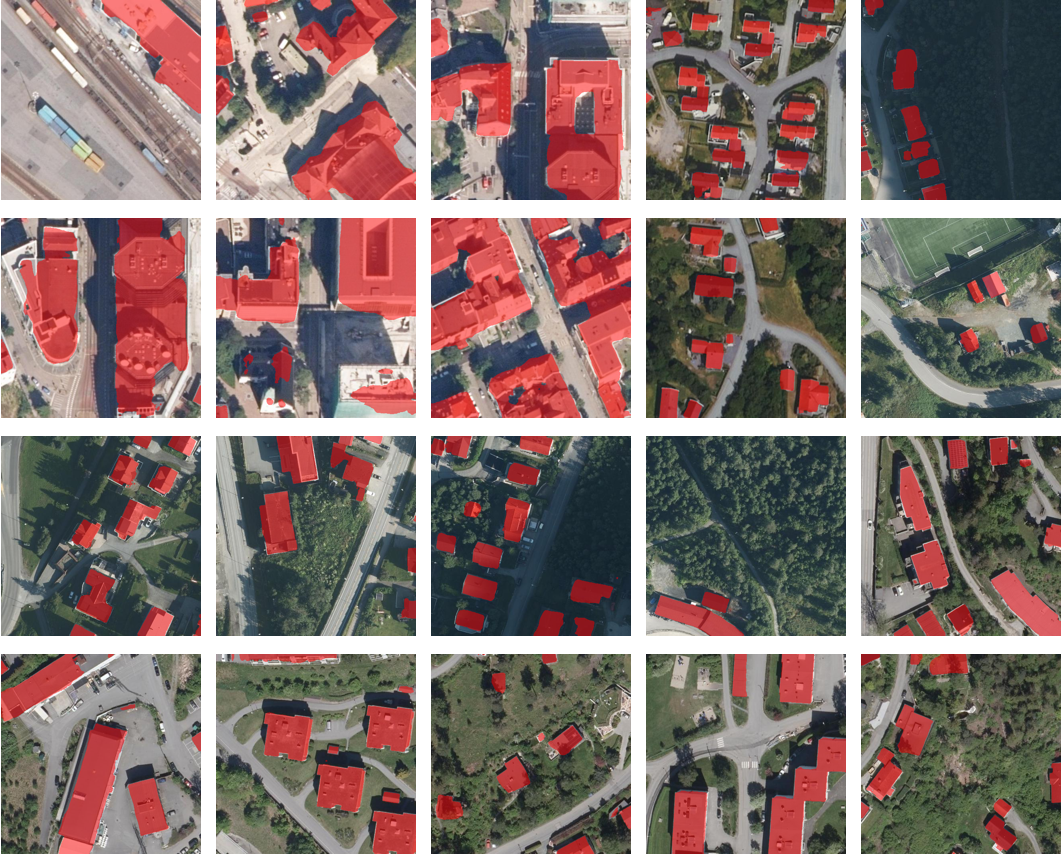}
	\end{center}
	\caption{The effectiveness of our pipeline in performing quantitative segmentation across various scenes and regions is presented in our study. Red mask is the building prediction.}
	\label{fig:result_vis}
\end{figure*}

Our study involved a thorough data analysis of the MapAI dataset, where we examined the impact of data fusion and pre-trained models on performance. Table \ref{tab:MapAI} presents compelling evidence of a substantial improvement in results when incorporating these techniques. Notably, in the fifth row of the table, a notable distinction is observed between SegFormer-B5 and SegAnyBuild, solely based on the inclusion or exclusion of pre-trained models. In our analysis, we established U-net, ConvNext, and SegFormer as baseline models for comparative purposes. The findings unequivocally demonstrate that our proposed framework outperforms these baseline models, thus establishing its superiority on the MapAI dataset. This achievement can be attributed to the synergistic effects of data fusion and the utilization of pre-trained models, showcasing the potential of our framework to advance the field of data analysis in the domains of computer vision and machine learning. The robust empirical evidence substantiates the effectiveness and competitiveness of our approach within the specific context of the MapAI dataset.

\begin{table}
    \centering
    \begin{tabular}{lcc}
        \hline Model & IOU & BIOU \\
        \hline U-Net & 0.7611 & 0.5823 \\
        ConvNext & 0.7841 & 0.6105 \\
        SegFormer-B0 & 0.7632 & 0.5901 \\
        SegFormer-B4 & 0.7844 & 0.6116 \\
        SegFormer-B5 & 0.7902 & 0.6185 \\
        \hline
        \hline
        SegAnyBuild & 0.8012 & 0.6213 \\
        \hline
        
    \end{tabular}
    \caption{Performance of different models on the MapAIcompetition image test set (without post-processing). As baseline we show a standard U-Net~\cite{long2015fully}, ConvNext~\cite{liu2022convnet}, SegFormer~\cite{xie2021segformer}. And Last row is our SegAnyBuild performance.}
    \label{tab:MapAI}
\end{table}

Furthermore, we conducted a comprehensive quantitative analysis by testing the dataset in various locations. The results, as depicted in Figure ~\ref{fig:result_vis}, clearly illustrate the effectiveness of our approach in performing building recognition and segmentation across diverse scenes in both urban and rural areas. This quantitative analysis further reinforces the robustness and generalizability of our framework in different geographical contexts. The ability to accurately identify and segment buildings in various settings, including towns and villages, highlights the versatility and practical applicability of our proposed methodology. These findings contribute to the growing body of evidence supporting the efficacy of our approach in addressing real-world challenges in the field of building recognition and segmentation.

\begin{table}
    \centering
    \begin{tabular}{lcc}
        \hline Dataset & IOU & BIOU \\
        \hline
        MapAI & 0.8012 & 0.6213 \\

        INRIA Aerial Image & 0.8265 & 0.6424 \\
        WHU Building  & 0.8452 & 0.6165 \\
        Floodnet & 0.5031 & 0.4012 \\
        \hline
        
    \end{tabular}
    \caption{Performance of SegAnyBuild model on the different Dataset.}
    \label{tab:allData}
\end{table}

Our investigation also encompassed testing our approach on various datasets, as demonstrated in Table ~\ref{tab:allData}. The results unequivocally substantiate the efficacy of joint training through data fusion, coupled with the utilization of pre-trained large models. Notably, our approach consistently achieved favorable outcomes across multiple datasets, thereby showcasing its robustness and generalizability. This ability to achieve impressive results on diverse datasets using a single model highlights the superiority and practicality of our proposed methodology. These findings contribute to the body of knowledge, providing empirical evidence of the effectiveness of our approach in addressing the challenges associated with multiple datasets, while emphasizing its potential for broader applications in the field.

\subsection{Ablation Study}
In Table ~\ref{tab:ablation}. Through our verification process, we have confirmed the feasibility of joint learning by fusing multiple building datasets. This observation arises from the understanding that buildings often exhibit similar texture features across various datasets. Consequently, there is significant potential for enhancing performance by learning deep features that capture these similarities. The fusion of multiple building data facilitates the extraction of shared features, enabling the model to leverage the collective knowledge embedded in different datasets. This approach offers a promising avenue for improving performance in building recognition and segmentation tasks. The results of our study provide empirical evidence supporting the notion that joint learning, driven by the fusion of multiple datasets, can effectively enhance performance by leveraging deep, similar features. These findings contribute to advancing our understanding of how to leverage shared knowledge across datasets for improved performance in building-related tasks.
\begin{table}
\begin{center}
\begin{tabular}{|l|cc|c|c|}
\hline
  Method  & self & fusion  & $IOU$($\uparrow$) & BIOU($\uparrow$) \\
\hline
SegAnyBuild-self   & \checkmark &   & 0.7642 & 0.6024\\
SegAnyBuild-fusion &  & \checkmark & 0.8012 & 0.6213 \\

\hline
\end{tabular}
\end{center}
\caption{The result is for using self data and fusion data on MapAI dataset.}
\label{tab:ablation}
\end{table}

\begin{figure}
    \centering
    \includegraphics[width=\columnwidth]{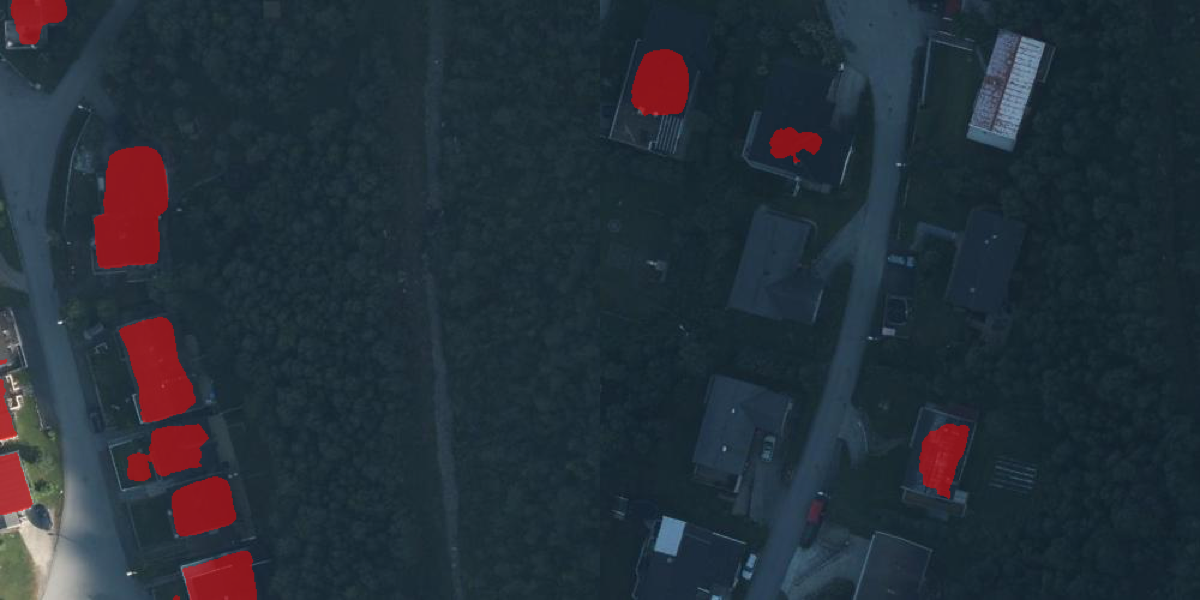}
    \caption{The segmentation results illustrate two distinct night scenes: on the left, a scenario featuring illuminated street lights, and on the right, a comparatively darker setting devoid of prominent light sources.}
    \label{fig:hard_case}
\end{figure}

\subsection{Discussion}
\paragraph{Diversification of Data Sources.} Our research successfully implemented robust building segmentation across an array of datasets, leveraging both dataset fusion techniques and prompts from pre-existing models. It is, however, crucial to recognize the availability of several other building datasets. When assimilated, these datasets could conceivably bolster building segmentation performance across a multitude of scenarios—especially under demanding circumstances such as low-lit scenes, inclement weather, and low-resolution imagery. As depicted in Figure ~\ref{fig:hard_case}, segmentation efficacy wanes in situations marked by intricate architectural nuances or inadequate lighting, such as nocturnal settings. This attenuation can largely be ascribed to a duo of causative factors. First, the restrictive nature of the extant data can circumscribe the segmentation prowess in the aforementioned arduous scenarios. Second, discerning the texture features of edifices becomes increasingly onerous in dimly lit or blurry settings, given the attenuated visibility and definition of salient characteristics. Such intricacies accentuate the challenges in executing segmentation under suboptimal conditions, thereby underlining the imperative for avant-garde models adept at navigating these constraints.

\paragraph{Incorporation of 3D Knowledge.} It is worth noting that specific datasets, such as the MapAI dataset, proffer invaluable depth information. Harnessing this depth data can amplify the segmentation process, providing additional cues that facilitate a more precise and consistent demarcation of edifice peripheries.

\paragraph{Advancements in Self-supervised Learning.} Delving deeper into the amalgamation of diverse datasets within the ambit of self-supervised learning ~\cite{zhou4425635multi, he2022masked, zhang2020fact, rahnemoonfar2021floodnet}, especially those endowed with depth intelligence, holds considerable promise. This strategy portends significant enhancements in building segmentation capabilities across a wider spectrum of scenarios. Such revelations serve as a linchpin in the evolving tapestry of research in this domain, emphasizing the latent potential of integrating supplementary datasets and depth insights to refine building segmentation methodologies.

\subsection{Conclusion}
The present research underscores the indispensable role of datasets culled from diverse sources and the utilization of advanced representation learning models, particularly in building segmentation within remote sensing imagery. Our methodological approach, characterized by the strategic fusion of several datasets, not only augments the available informational spectrum for learning but also showcases superior performance across the entire gamut of datasets harnessed. The triumphant realization of a unified training paradigm stands testament to the robustness of our strategy, potentially ushering in transformative advancements in pivotal sectors such as urban planning, disaster mitigation, and environmental monitoring. Our pioneering integration of dataset fusion methodologies and insights from pre-existing models marks a distinct shift from traditional approaches, thereby redefining the modus operandi of building segmentation challenges within the domain of remote sensing imagery. This research, in essence, lays a solid groundwork for subsequent explorations, opening vistas for potentially groundbreaking innovations and applications in the realm of building segmentation.

% Note: If you are using BibTeX, please use the following code:
% \bibliographystyle{spbasic}
\bibliography{ref} 
% where "bib_CGIconf" has to be replaced by the name of your BibTeX
% file (without the .bib extension). 

% Non-BibTeX users please use:
% \begin{thebibliography}{}
% % and use \bibitem to create references as below. For the names of
% % the authors, please write each name followed by the initials,
% % e.g.: Knuth D.E., Lamport L.
% \bibitem{RefA}
% Author, Article title, Journal, Volume, page numbers (year)
% \bibitem{RefB}
% Author, Book title, page numbers. Publisher, place (year)
% \end{thebibliography}
% \end{sloppypar}

\end{document}